# LiteMORT: A memory efficient gradient boosting tree system on adaptive compact distributions

Yingshi Chen

[1] Institute of Electromagnetics and Acoustics, and Department of Electronic Science, Xiamen University, Xiamen 361005, China
E-mail: gsp@grusoft.com

**Abstract**

Gradient boosted decision trees (GBDT) is the leading algorithm for many commercial and academic data applications. We give deep analysis of this algorithm, especial the histogram technique, which is a basis for the regulized distribution with compact support. We present three new modifications. 1) Share memory technique to reduce the memory usage. In many cases, only need the data source itself and no extra memory. 2) Implicit merging for "merge overflow problem". "merge overflow" means that merge some small datasets to a huge datasets, which are too huge to be solved. By implicit merging, we just need the original small datasets to train the GBDT model. 3) Adaptive resize algorithm of histogram bins to improve the accuracy. Experiments on two large kaggle competitions verified our methods. They use much less memory than LightGBM and have higher accuracy. We have implemented these algorithms in an open source package LiteMORT. The source codes are available at https://github.com/closest-git/LiteMORT

**1. Introduction**

Gradient boosting algorithm [1] is one of the most interesting and overlooked algorithm in machine learning. There are huge gaps between the simple theoretical formula [1] and practical implementations [2,3,4,5]. In particular, histogram-based gradient boosted trees [3] is the best algorithm combination for many problems. The histogram technique not only greatly improves the speed, but also improves the accuracy. In many real applications and most Kaggle tabular-data competitions, histogram-based lightGBM always has higher accuracy than XGBoost, CatBoost and any other non-GBDT tools. It's clear that histogram technology plays an important role. But There are still no clear theory to explain the relation between the histogram technique and the accuracy. In this paper, we think histogram representation is actually a basis for the regulized distribution with compact support. Along this direction, we give some improvements to reduce memory drastically and improve the accuracy further. We have implemented these algorithm in LiteMORT, which is an open-source Python/C++ package

The first purpose of histogram-based algorithm is to reduce the time to find the best split, which is the time bottleneck in the previous implementation. For a dataset(or subset) with $N$ samples $\mathbf{X} = \{x_i\}$ and $M$ features $\{\phi_j\}$. A $K$ bin histogram would reduce the tree-split algorithm complexity from O($NM$) to O($KM$). The common value of $K$ is 256 (or 64,512,...), which is much smaller than N. So this reduces the time to find the best split drastically.

Why this form would improve the accuracy? In the process of gradient boosting, we need to calculate many variables, for example, gradient, hessian, newton direction… Let $\sigma$ is the distribution parameters(mean, devian, sum of squares, etc) for each bin's variable. Then as formula (1), we use these $\sigma$ as a new distribution to represent the original features $\phi$. Since the statistical parameters of each bin are much more robust than single value, $H(\phi)$ would provide more robust and generative information for machine learning task. Once upon an element is put into some bin, its own feature value is not important. In all related operation, we only use $\sigma$ value of its bin. In some sense, the statistic value of one bin is a regulation to all single values in this bin. Just like a smooth filter to reduce the noise. So make the decision more generative.

$$\boldsymbol{Distribution}: \ H(\boldsymbol{\phi}) = \{\sigma(bin_1), \sigma(bin_2), \cdots, \sigma(bin_K)\} \qquad (1)$$

$$where \ \{x_1, x_2, \cdots, x_N\} = bin_1 \cup bin_2 \cup \cdots \cup bin_K$$

We think that $H(\phi)$ is actually a regulized distribution with compact support. And we could get more along this direction. In this paper, we present three improvements.

**1) feature distribution on adpative resized bins**. The histogram bins is actually a basis to calculate distribution of different variables. With more refined bins, we could get better distribution parameters, which would have better generality. Even if we should fix the size of bin to 256, we can adjust the size of each bin. We find some metric to estimate each bin's importance. If some bin is more important, then we will shrink its size or split to two bins. Then in the later training, we could get a better split point.

**2) share memory with data source**. In practical application, the main bottleneck comes from the huge memory requirements. The size of huge tabular datasets may be several hundred G, several thousand G, or even more. The most widely used software, such as XGBoost, LightGBM or CatBOOST would load these datasets into memory and need more memory to process these dataset. The amount of memory required is usually 2-3 multiple of the training set size[7]. As indicated in the above analysis, What is really GBDT needed is the distribution value at each bin. So LiteMORT shares memory with data source. The amount of memory required is usually just the training set.

**3) implicit merging.** Merging data from different data sources is one of the most common tasks that data scientists have to do. In most case, the merged features have many duplicate items. Especial for the feature generated from groupby operation. In this common groupby/merge operation, we would get a huge column with 10 million items, but has only several unique items. And more, the merged datasets are too huge to be solved by GBDT. We call this problem as "merge overflow problem". We provide a novel implicit merging technique for this problem. That is, no need to merge many small datasets to a huge datasets. Just send these small datasets to LiteMORT, then LiteMORT would automatically generate the histogram structure and all other distributions information needed in the training process. Then LiteMORT would finish the training with same accuracy, at the same time, need only one tenth, one percent or even less memory of standard merge operation.

We have implemented these algorithms in LiteMORT, which is an open-source Python/C++ package and the codes are available at https://github.com/closest-git/LiteMORT. LiteMORT uses much less memory than LightGBM and other GBDT libs. Experiments show that it's much faster than LightGBM[11] with higher accuracy at some datasets. LiteMORT reveals that GBDT algorithm can have much more potential than most people would expect.

## 2. Methodology

To describe the problem more concisely, we give the following formulas and symbols:

For a dataset with N samples $\mathbf{X} = \{x_i\}$ and M features $\{\phi_j\}$, the Gradient Boosting algorithm would learn a series of additive functions $\{f^1, f^2, f^3, ...\}$ to predict the output:

$$\hat{y}_i = \sum f_i^t(\mathbf{X}) \quad f^t, h^t \in \mathcal{F} \quad (2)$$

$\mathcal{F} = \{f\}$ is a functional space including all kinds of learning functions (decision tree, convolution network ). The boosting framework can be generalized to any choice of base learner, but nearly all popular GBDT lib use decision trees because they work well in practice. LiteMORT would try more base learners.

In the training process, we use loss function $l(y_i, \hat{y}_i)$ for the difference between the predicted value $\hat{y}_i$ and the target value $y_i$. The $\hat{y}_i$ at current step $t$ would be derived from the previous step $(t-1)$

$$\mathcal{L}^{(t)} = \sum_{i=1}^{N} l\left(y_i, \hat{y}_i^{(t-1)} + cf_i^t(\mathbf{X})\right) \quad (3)$$

In the following derivation, the superscript $t$ can be omitted. That is, if some variable has no superscript, it always represents the value at current step $t$. Ignore the higher order residuals, the functional (3) is further expanded as follows

$$\mathcal{L}^{(t)} \approx \sum_{i=1}^{n} l\left(y_i, \hat{y}_i^{(t-1)}\right) + g_i f(x_i) + \frac{1}{2} h_i f^2(x_i) \quad (4)$$

where $g_i, h_i$ is the first and second derivative of loss function at $\hat{y}_i^{(t-1)}$:

$$g_i = \frac{\partial l(y_i, \hat{y}_i^{(t-1)})}{\partial \hat{y}_i^{(t-1)}} \qquad h_i = \frac{\partial^2 l(y_i, \hat{y}_i^{(t-1)})}{\partial^2 \hat{y}_i^{(t-1)}} \tag{5}$$

Since at step $t$, $l(y_i, \hat{y}_i^{(t-1)})$ is constant, the second-order functional (4) is further simplified to $G^{(t)}$:

$$G^{(t)} \approx \sum_{i=1}^{n} g_i \tilde{f}_t(x_i) + \frac{1}{2} h_i \tilde{f}_t^{\,2}(x_i) \tag{6}$$

(6) is a gereal functional for all base funcitons. We would give some simple but important results for the case of decision trees. A decision tree with T leafs is generated by series of split on some feature $\{\phi_1, \phi_2, \cdots, \phi_{T-1}\}$. We call $\{\phi_1, \phi_2, \cdots, \phi_{T-1}\}$ as the basis feature of this tree. Decision trees have two unique features: 1) Every sample $x_i$ would locate at some leaf $j$. We use leaf mapping $I_j = \{i | leaf(x_i) = j\}$ to represent this relation. 2) Each leaf $j$ has a unique value $\omega_j$, and all samples in this leaf would have same value $\omega_j$! So we expand the general functional (6) with tree's leaf mapping and leaf value:

$$G^{(t)} \approx \sum_{i=1}^{n} g_i \tilde{f}_t(x_i) + \frac{1}{2} h_i \tilde{f}_t^{\,2}(x_i) = \sum_{j=1}^{T} \left[ \left(\sum_{i \in I_j} g_i\right) \omega_j + \frac{1}{2} \left(\sum_{i \in I_j} h_i\right) \omega_j^2 \right] \tag{7}$$

At each leaf, let $g_J = \sum_{i \in I_j} g_i \quad h_J = \sum_{i \in I_j} h_i$

$$G^{(t)} \approx \sum_{j=1}^{T} \left[ g_J \omega_j + \frac{1}{2} h_J \omega_j^2 \right] = \sum_{j=1}^{T} \left[ \frac{h_J}{2} \left( \omega_j + \frac{g_J}{h_J} \right)^2 - \frac{(g_J)^2}{2 h_J} \right] \tag{8}$$

It's easy to see to get the extream value, $\omega_j = -\frac{\sum_{i \in I_j} g_i}{\sum_{i \in I_j} h_i}$ ,which is actually the newton direction.

And the extream value is:

$$G^{(t)} = -\frac{1}{2} \sum_{j=1}^{T} \frac{\left(\sum_{i \in I_j} g_i\right)^2}{\sum_{i \in I_j} h_i} \tag{9}$$

In the case of simplest MSE(Mean Square Error) loss function:

$$l(y_i, \hat{y}_i) = \frac{1}{2}(y_i - \hat{y}_i)^2 \tag{10}$$

$h_i = 1$, So at each leaf, the $\omega_j$ is just the average of $g_i$. That is

$$\omega_j = -\frac{\sum_{i \in I_j} g_i}{\sum_{i \in I_j} h_i} = \frac{\sum_{i \in I_j} g_i}{|I_j|} \tag{11}$$

So the mean value of all leafs is actually the newton direction of MSE loss function.

**Share memory with data source**

As described above, each decision tree is generated by some basis features $\{\phi_1, \phi_2, \cdots, \phi_{T-1}\}$. In practical application, each $\phi$ has many different formats and data structures. For example, the dataframe of pandas, ndarray of numpy, list, vecotr…Anyway, these features are always stored in some place of memory. In most case, all values of one feature are stored continuously. So we only need the pointer of first element. All other elements can be accessed by offset, which is just the length of its data type multiplied by its index. In the implementation of LiteMORT, nearly all visit to $\{\phi_1, \phi_2, \cdots, \phi_{T-1}\}$ would be replaced by the feature pointer $\{p(\phi_1), p(\phi_2), \cdots, p(\phi_{T-1})\}$. With this share memory technique, LiteMORT would not allocate extra memory for features stored in continuous memory. In the gradient boosting process, nearly all visit to data is on the pointer and some offsets. So no extra memory needed.

**Implicit merging for "merge overflow problem"**

In real application, we usually don't save all the data in one big data table. They are always many smaller ones instead. But in the data analysis or machine learning task, we have to access all datas. Or we have to merge some small datasets to get some huge datasets, which are too huge to be processed by many classical machine learning algorithms. We called this phenomenon as "merge overflow problem". Not long ago, GBDT was also encountered many difficulties to deal with such "merge overflow problem". But If we carefully analyze the algorithm of GBDT, as the formula (2)-(9), what GBDT really needs is some compact distributions, not all values of a huge features. And in the case of "merge overflow problem", it's easy to get the distribution from all its source, no need to actually generate a huge matrix. So as figure 1 shows, just send small datasets from the source to LiteMORT, then LiteMORT would generate the histograms for each merged features. In the later training process, all operations are on these histograms. No need to generate the huge merged dataset as the classical method. So implicit merging algorithm shows another potential of GBDT algorithm , which can handle the merge-dataset problem by a memory-efficient way. While many other machine learning algorithms have to use explicit merge operation and may fail because of memory limitation.

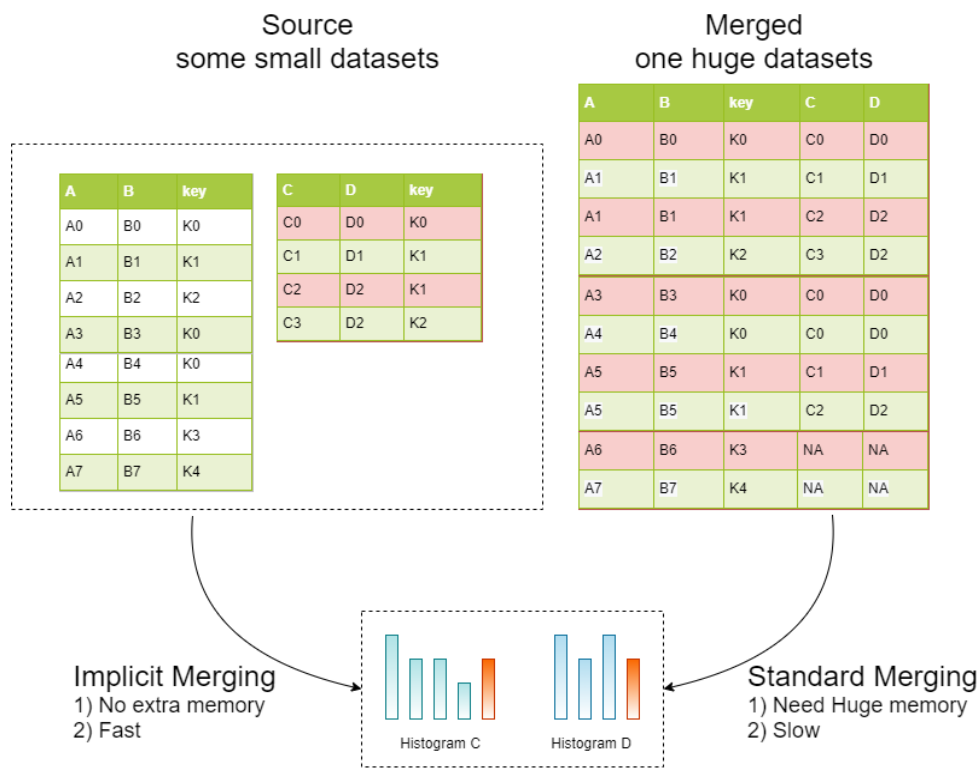

Figure 1 Implicit merging algorithm for "merge overflow problem"

We should point that the actual implemenation is much more complex and harder than the simple descreption. There is a huge gap between its simple appearance and realization,which do need hard working and coding.

**Adaptive Distribution on adaptive resize of histogram bins**

To improve the speed to find the best split point, each feature is represented by a histogram with limited size of bins. The possible split points are also limited since they are always at the border of bins. We use adaptive resize algorithm to find the split at more positions. That is, even the number of bins is still fixed, their sizes are dynamic changed in the training process. As algorithm 1 shows, we use a metric to evalute the importance of each bins, then modified some important bins in the training process. With this adaptive resize process, we could get better split of bins .As the training going on, the split position would be more precise.

Algorithm 1 Adaptive resize of histogram bins in the training process

> Set $T_{split}$ to some value ( for example 32)
> Init the histogram bin before the training
> At each step in the training
>   If some bin $b_k$ of some featue $\phi_j$ is the split point, then
>   $$N(b_k) = N(b_k) + 1$$
>   If $N(b_k) > T_{split}$:
>     Refine the bins of feature $\phi_j$:
>     1) If the number bins of histogram less than 256, divide the bin $b_k$ to two bins
>     2) Else, reduce size of $b_k$. Other bins in this histogram would be enlarged.

## 3. Results and discussion

To verify our method, we test its performance on two latest kaggle competitions. The first competition is "IEEE-CIS Fraud Detection" (**https://www.kaggle.com/c/ieee-fraud-detection**). We use 8-fold cross validation in the training process. Table 1 and table 2 compare its performace with LightGBM. In each table, the row named "LightGBM" is corresponding to the result of latest LightGBM 2.2.3; the row named "LiteMORT" is corresponding to the result of LiteMORT 0.1.18(without adaptive distribution); the row named "LiteMORT-adaptive" is corresponding to the result of LiteMORT with adaptive distribution. We can see that :

1) In this competition, LiteMORT is much faster than LightGBM. LiteMORT needs only a quarter of the time of lightGBM.

2) In this competition, LiteMORT has higher auc than lightGBM. And with adaptive algorithms, "liteMORT-adaptive" would achieve higher accuracy.

Table 1   AUC of each fold

| fold | 1 | 2 | 3 | 4 | 5 | 6 | 7 | 8 |
|---|---|---|---|---|---|---|---|---|
| LightGBM | 0.9802 | 0.9813 | 0.9794 | 0.9787 | 0.979 | 0.9787 | 0.9775 | 0.9758 |
| LiteMORT | 0.9807 | 0.983 | 0.9793 | 0.9789 | 0.9799 | 0.9786 | 0.977 | 0.9778 |
| LiteMORT-adaptive | 0.9816 | 0.9831 | 0.98 | 0.9805 | 0.9803 | 0.9801 | 0.9785 | 0.9791 |

Table 2   Time of each fold

| fold | 1 | 2 | 3 | 4 | 5 | 6 | 7 | 8 |
|---|---|---|---|---|---|---|---|---|
| lightGBM | 3184 | 3212 | 2869 | 2866 | 2794 | 3179 | 2564 | 2253 |
| liteMORT | 608.2 | 657.1 | 513.6 | 559.5 | 609.9 | 502.1 | 499.4 | 551.1 |
| liteMORT-adaptive | 630.2 | 665.1 | 661.4 | 623.3 | 629.9 | 675.1 | 631.1 | 655 |

The second competition is the "ASHRAE - Great Energy Predictor III" (**https://www.kaggle.com/c/ashrae-energy-prediction**). In this case (to merge 25 features). We use 5-fold cross validation in the training process. Table 3 compare its peak memory performace with LightGBM. In each table, the row named "LightGBM" is corresponding to the result of latest LightGBM 2.2.3, which need 28.9G memory. The row named "LiteMORT" is corresponding to the result of LiteMORT 0.1.18(without implicit merging), which need 14.2G memory. The row named "LiteMORT- Implicit merging" is corresponding to the result of LiteMORT with implicit merging, which only need about 4.5G memory. Table 4 lists the RMSE of each fold. We can see that

1) In this competition, LiteMORT use only 50% memroy of LightGBM. With implicit mergin technique, only need **16%** memroy**.**

2) In this competition, LiteMORT has same RMSE with lightGBM. This verified our memory efficient algorithm does not affect the accuracy of GBDT algorithm.

Table 3  Peak Memory Comparison

| method | Peak Memory |
|---|---|
| LightGBM | 28.9(G) |
| LiteMORT | 14.2(G) |
| LiteMORT-Implicit merging | 4.5(G) |

Table 4  RMSE of each 5-fold

| fold | 1 | 2 | 3 | 4 | 5 |
|---|---|---|---|---|---|
| lightGBM | 1.042 | 0.981 | 1.00 | 1.098 | 0.980 |
| liteMORT | 1.045 | 0.981 | 0.998 | 1.098 | 0.978 |
| liteMORT-adaptive | 1.036 | 0.963 | 0.989 | 1.090 | 0.968 |

We would see in both competitions, LiteMORT is much better than lightGBM. That's not to say that's true in all cases. As the famouse no free lunch theorem, some lib would perform better in some datasets and maybe poor in other datasets. These two competions shows the potential of LiteMORT and the potential of GBDT. Although the best GBDT libraries(LightGBM, CatBoost, LiteMORT) have made great progress, at least they have defeated other algorithms in nearly all tabular data applications. We are sure it is only halfway, far from the end.

## 4 Conclusion and Prospect

LiteMORT is a new GBDT library on on adaptive compact distributions. With implicit merging and share memory technique, LiteMORT uses much less memory thant LightGBM and other similar tools. LiteMORT maybe the first GBDT libs that solved the "merge overflow problem". With adaptive resize of histogram bins LiteMORT would get higher accuracy on some datasets. So it's a very valuable tool for various academic research and business applications. We will continue work hard to improve its speed,memory usage and accuracy.